%% file: neurips_2026.tex
\documentclass{article}

    \PassOptionsToPackage{numbers, compress}{natbib}

\usepackage{booktabs}
\usepackage{makecell}
\usepackage{graphicx}
\usepackage[preprint]{neurips_2026}

\usepackage[utf8]{inputenc} 
\usepackage[T1]{fontenc}    
\usepackage{hyperref}       
\usepackage{url}            
\usepackage{amsfonts}       
\usepackage{nicefrac}       
\usepackage{microtype}      
\usepackage{xcolor}         

\usepackage{amsmath} 
\usepackage{array}
\title{OmniShotCut: Holistic Relational Shot Boundary Detection with Shot-Query Transformer}

\author{%
Boyang Wang$^{1}$,
Guangyi Xu$^{1}$,
Jiahui Zhang$^{1}$,
Zhipeng Tang$^{2}$,
Zezhou Cheng$^{1}$
\\
$^{1}$University of Virginia \quad
$^{2}$University of Massachusetts Amherst
\\[0.9em]
{\small
\textbf{Project Page}: 
\href{https://uva-computer-vision-lab.github.io/OmniShotCut_website/}
{\textcolor{blue}{\texttt{Omni-Shot-Cut.github.io}}}
}
}

\begin{document}

\maketitle
\vspace{-2.0em}

\input{sec/0_abstract}

\input{sec/1_intro}

\input{sec/2_related_work}

\input{sec/3_method}

\input{sec/4_experiment}

\input{sec/5_conclusion}

\bibliographystyle{plainnat}
\bibliography{references}

\appendix

\clearpage

\input{sec/6_supp}

\clearpage


\end{document}

%% file: sec/0_abstract.tex
\begin{abstract}
Shot Boundary Detection (SBD) aims to automatically identify shot changes and divide a video into coherent shots.
While SBD was widely studied in the literature, existing methods often produce non-interpretable boundaries on transitions, miss subtle yet harmful discontinuities, and rely on noisy, low-diversity annotations and outdated benchmarks. 
To alleviate these limitations, we propose \textbf{OmniShotCut} to formulate SBD as structured relational prediction, jointly estimating shot ranges with \textit{intra}-shot relations and \textit{inter}-shot relations, by a shot query-based dense video Transformer. 
To avoid imprecise manual labeling, we adopt a fully synthetic transition synthesis pipeline that automatically reproduces major transition families with precise boundaries and parameterized variants. 
We also introduce OmniShotCutBench, a modern wide-domain benchmark enabling holistic and diagnostic evaluation.
Experiments on the benchmarks demonstrate the effectiveness and generality of our method.
\end{abstract}


%% file: sec/1_intro.tex
\section{Introduction}

Modern video production is inherently compositional, where multiple shots are assembled through editing operations rather than captured in a single continuous take.
The transitions between these shots follow artistic principles, spanning from abrupt hard cuts and jump cuts to gradual effects such as dissolves, fades, wipes, etc. 
To understand the structural composition of such edited videos, it is necessary to identify the most atomic temporal units, a group of frames that form a coherent shot. This task is known as Shot Boundary Detection (SBD).

Shot Boundary Detection~\cite{zhu2023autoshot, soucek2024transnet, souvcek2019transnet, tang2018fast} has long been regarded as a well-established problem in video understanding. 
However, despite its apparent maturity, progress in this area has stagnated. 
We revisit SBD from the perspective of its downstream applications and ask: has the problem truly been well defined and solved, and is it addressed in the most efficient and scalable manner?
We argue that current SBD pipelines remain limited along several practical axes.

\begin{figure}[t]
    \centering
    \vspace{-0.3cm}
    \includegraphics[width=\textwidth]{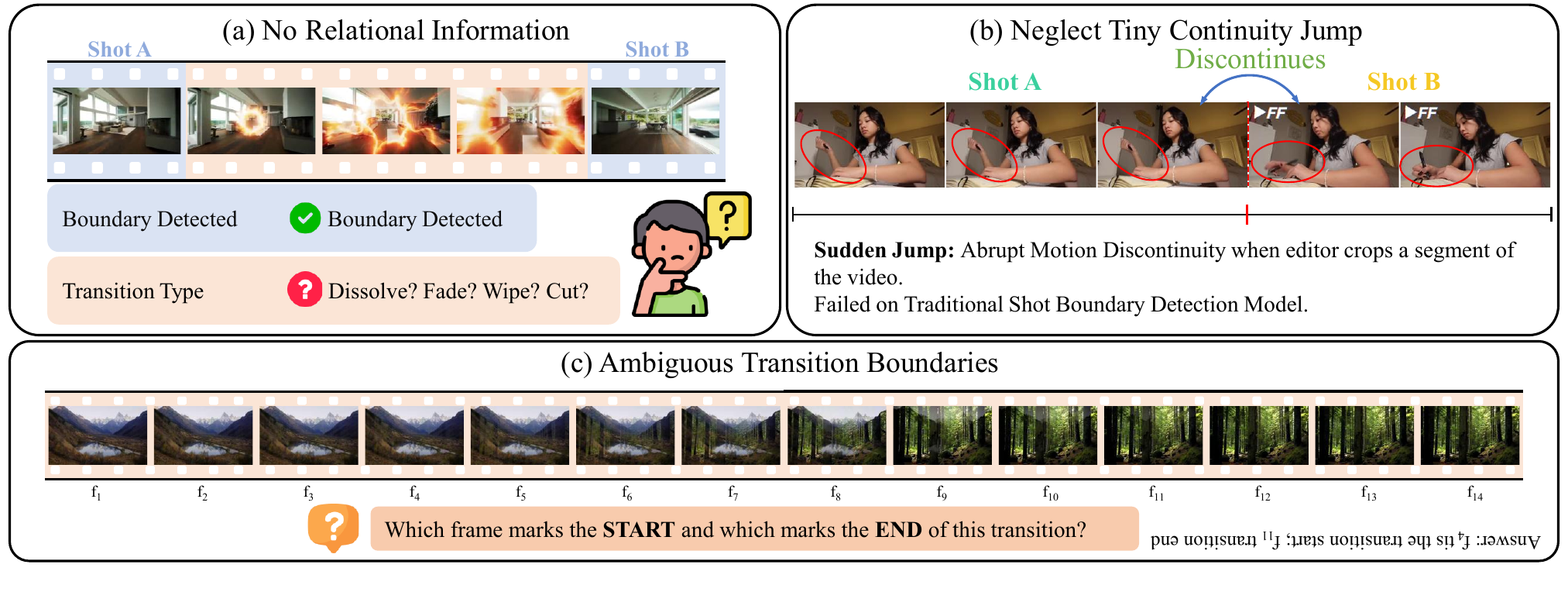}
    \vspace{-1.0cm}
    \caption{
    \textbf{Limitations of traditional Shot Boundary Detection models.}
        (a) Detected Shots are hard to interpret: predicted boundaries lack explicit transition semantics;
        (b) Sudden Jump is under-modeled and often missed;
        (c) Human annotations are unreliable for gradual transitions with subtle start/end frames.
    }
    \label{fig:teaser}
\end{figure}

First, the predicted shots lack interpretability, as it is unclear whether a predicted boundary corresponds to a scene or an editing transition (see Fig.~\ref{fig:teaser}a).
For each detected shot, the output should not be limited to a simple temporal range, but should also include higher-level structural information that better supports downstream applications.
For instance, in video generation~\cite{wan2025wan, yang2024cogvideox, agarwal2025cosmos}, transition may be less critical, and clean vanilla shot segments are often preferred. 
To this end, we introduce \textbf{\textit{intra}-shot} relation classification as the outputs of the model. 
The intra-label characterizes the shot itself, indicating whether it is a vanilla segment or a specific transition type.

Second, previous SBD models fail to detect subtle yet harmful discontinuities (i.e., sudden jumps) that negatively affect downstream tasks.
Sudden jump (see Fig.~\ref{fig:teaser}b) introduces excessive abrupt motion or texture change in two consecutive frames, which exert negative influence on motion tracking~\cite{karaev2025cotracker3}, video segmentation~\cite{hu2025segment}, latent video compression~\cite{agarwal2025cosmos, yang2024cogvideox, wan2025wan}, and more downstream tasks.
To this end, we introduce \textbf{\textit{inter}-shot} relation classification.
The inter-label captures its relationship with the preceding shot, modeling the cross-shot continuity relationship.
Further, existing state-of-the-art SBD models, such as TransNetV2~\cite{soucek2024transnet} and AutoShot~\cite{zhu2023autoshot}, rely on 3D CNN architectures that are not well-suited for our richer formulation. 
Instead, we design a shot query-based Transformer architecture that jointly optimizes all objectives through shared hidden states, enabling unified modeling of temporal shot range prediction and relational understanding.

\begin{figure}[t]
    \centering
    \vspace{-0.3cm}
    \includegraphics[width=\textwidth]{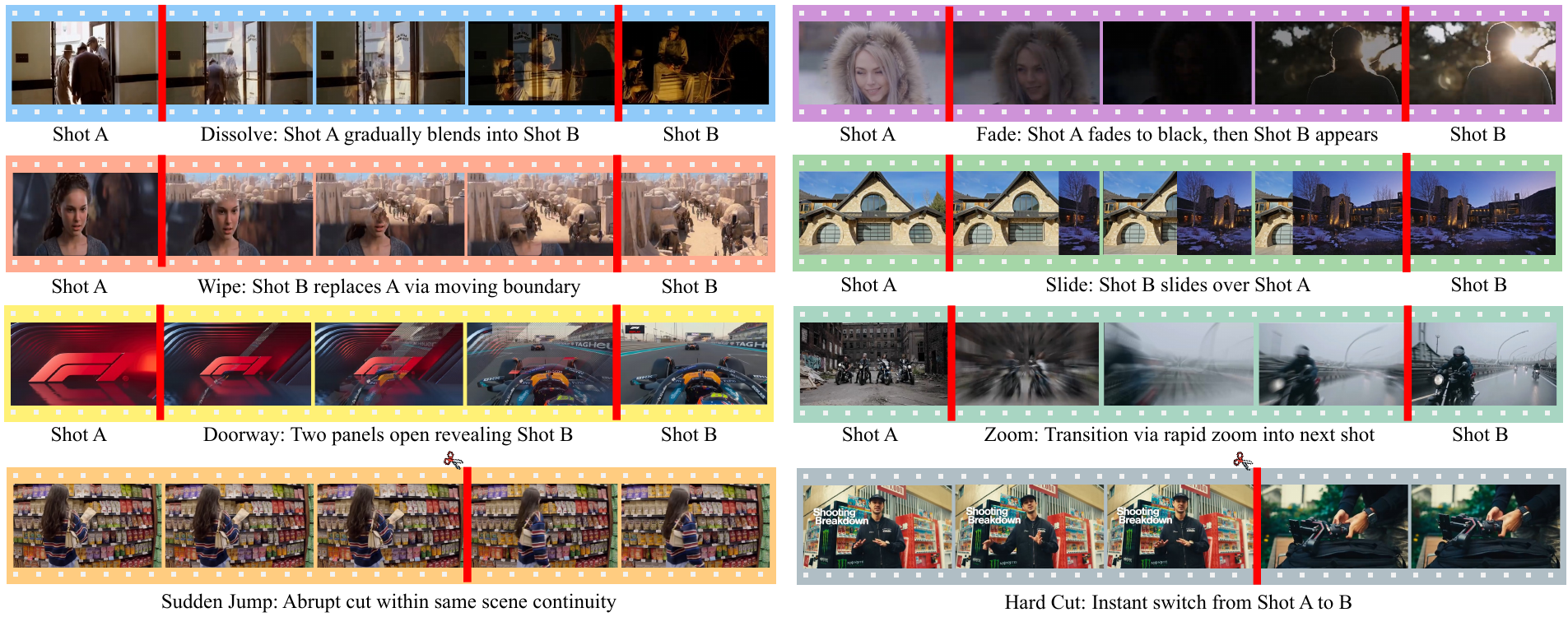}
    \caption{
       \textbf{Main Transition Types.}
       We consider a diverse and comprehensive set of video shot transitions that are largely underexplored in prior shot boundary detection works~\cite{zhu2023autoshot,soucek2024transnet,souvcek2019transnet}. This figure illustrates several representative transition types, including dissolve, fade, wipe, slide, doorway, zoom, sudden jump, and hard cut. 
       Each example shows the temporal progression from Shot A to Shot B with the transition region highlighted. 
       We skip the Pushing effect demo here.
       More subtypes are provided in the supplementary.
    }
    \label{fig:transition_types}
\end{figure}

Third, human-supervised annotation struggles to localize subtle changes accurately (see Fig.~\ref{fig:teaser}c).
Prior works~\cite{baraldi2015deep, baraldi2015shot, awad2017trecvid} heavily rely on manually annotated real-world data for shot boundary detection. However, transition labeling is highly labor-intensive and inherently imprecise. In particular, humans struggle to accurately localize subtle boundaries, such as the exact start and end frames of fading or dissolve effects, where minor changes in illumination and transparency are difficult to perceive. 
As a result, manual annotation is not well-suited for fine-grained transition modeling.
More importantly, transition effects are in fact generated by video editing software (e.g., Apple iMovie or Adobe professional editing suites). 
Instead of investing costly human effort in reverse annotation, we propose a forward generation strategy that programmatically reproduces transitions (Fig.~\ref{fig:transition_types}), which covers 9 main types and 30 subtypes, and yields hundreds of variations by sweeping controllable parameters in directions, edges, intensity, layout, and more.
This methodology enables the construction of a synthetic training dataset with precise transition ranges, while covering rare yet realistic cases (e.g., mosaic, puzzle, cube, doorway) that are underrepresented in existing datasets.
Furthermore, naively stitching together unrelated videos does not reflect real-world editing patterns. 
To address this, we leverage a self-supervised learning method to group semantically similar videos from our million-scale video clips pool, thereby simulating more realistic transition contexts. 

Fourth, existing benchmarks contain noisy annotations, outdated and narrow domain video sources, and miss the focus of the sudden jump, which fail to reflect the diversity and complexity of modern video content.
To close this gap, we introduce \textbf{OmniShotCutBench}, a contemporary SBD benchmark built from wide-domain, highly complex sources, with intra- and inter-shot relational labels and, importantly, a confidence scoring system to account for human judgment uncertainty.
We hope OmniShotCutBench can offer a more holistic and diagnostic evaluation for modern Shot Boundary Detection.

In summary, our contributions are as follows:
\begin{itemize}
    \item We reformulate Shot Boundary Detection by enriching each shot with both intra-shot and inter-shot relational information, moving beyond simple temporal range prediction.
    \item We propose a shot query-based Transformer architecture that jointly optimizes range prediction and relational classification within a unified hidden state.
    \item We introduce a fully synthetic pipeline with novel self-supervised clustering that groups similar but not identical clips and provides precise, diverse labels without manual annotation.
    \item We introduce a new benchmark for modern shot boundary detection that captures diverse contemporary transition patterns and sudden jumps, where our model consistently outperforms existing methods across multiple evaluation dimensions.
\end{itemize}

%% file: sec/2_related_work.tex
\section{Related Works}

\paragraph{Shot Boundary Detection.}
Shot Boundary Detection~\cite{kar2024video} operates on native, full video sequence inputs without frame downsampling. It requires frame-level precision to localize each boundary. 
This inherently demands high-density temporal inputs for the model. 
Traditional approaches, such as PySceneDetect~\cite{pyscenedetect2025} and Koala-36M~\cite{wang2025koala}, primarily rely on handcrafted low-level features (e.g., color histogram differences or structural similarity) to detect abrupt transitions. These methods are sensitive to illumination changes and often struggle to capture higher-level semantic consistency across frames.
Deep learning-based methods have since become dominant, including DeepSBD~\cite{hassanien2017large}, ClipShots~\cite{wang2019detecting}, TransNetV2~\cite{soucek2024transnet}, and AutoShot~\cite{zhu2023autoshot}. They employ 3D CNNs to detect transition intervals. 
To handle long sequences efficiently, these models often downsample spatial resolution aggressively (e.g., to $48 \times 27$) to reduce computational cost.
To evaluate SBD, even though benchmarks like BBC~\cite{baraldi2015deep}, RAI~\cite{baraldi2015shot}, and IACC3~\cite{awad2017trecvid} are popularly used with corresponding training datasets, many shot cut labels lack a clear definition and neglect transitions and subtle motion discontinuity cases like sudden jumps.
In addition, these datasets are primarily derived from legacy broadcast footage and do not reflect the diversity of modern video domains.

\paragraph{Downstream Applications.}
Shot Boundary Detection has become increasingly important in dataset curation for internet-scale in-the-wild videos. 
In data curation, massive raw long-form videos must be segmented into temporally coherent clips without abrupt changes. 
The temporal consistency of each clip is critical for training downstream models that demand continuous sequences of images, like the video generation task. 
In state-of-the-art video generation, videos are encoded by a temporal VAE~\cite{yang2024cogvideox, wan2025wan}, where multiple frames share the same token. 
If sudden jumps are not accurately detected, consecutive frames may exhibit drastic spatial shifts (e.g., a subject abruptly moving from left to right), significantly hindering temporal compression and modeling.
Moreover, shot boundary detection models are widely used to filter in-the-wild online videos and have been incorporated into numerous recent datasets and benchmark works~\cite{li2025sekai, lin2025towards} as the key component in curation preprocessing. 
In the scene segmentation task, MovieNet~\cite{huang2020movienet} provides a dataset that is first cropped by SBD, and then the following works, like LGSS~\cite{rao2020local}, BaSSL~\cite{mun2022bassl}, ShotCOL~\cite{chen2021shot}, and Scene-VLM~\cite{berman2025scene}, apply downsampled frames from the predicted shot boundaries to detect scene boundaries.
This growing reliance further underscores the need for a more precise, scalable, and transition-aware shot boundary detection model.

\paragraph{Synthetic Data.}
While supervised training on manually annotated pairs remains a standard approach in computer vision, several domains have noted the difficulty of collecting precisely aligned data in certain tasks. 
As a result, synthetic data generation strategies~\cite{mumuni2024survey} have been increasingly adopted, leveraging programmable forward transformations to automatically construct large-scale labeled datasets.
A representative example arises in low-level vision~\cite{wang2021real, jeelani2023expanding, wang2024vcisr}, where perfectly pixel-aligned degraded inputs and original high-quality pairs are rarely available in real-world settings. 
Instead, their degraded images or videos are synthesized via degradation to high-quality ground-truth sources. 
Similarly, prior research~\cite{wang2019detecting} in image forensics and editing detection has leveraged scripted pipelines to automatically synthesize Photoshop-manipulated images. 
For transition modeling, where effects are typically synthesized by editing software, a programmatic synthesis strategy is therefore a principled and effective solution.
Though previous works like TransnetV2~\cite{soucek2024transnet} and DeepSBD~\cite{hassanien2017large} mix real data with synthesized hard-cut and dissolve, most transition effects remain underexplored, and we extend to dozens of transitions available. 
More importantly, we explore the extent to which purely synthetic training data can push the limits of synthetic supervision.

%% file: sec/3_method.tex
\section{Method}
\paragraph{Problem Formulation.}
We define the problem as empowering traditional Shot Boundary Detection within an end-to-end model such that it not only predicts the temporal ranges of each shot, but also outputs the intra-relation classification of the shot itself and the inter-relation classification with respect to the previous shot.
For intra-relation, we include 8 major categories, which includes vanilla General video, Dissolve, Wipe, Push, Slide, Zoom, Fade, and Doorway. 
For inter-relation, we classify whether the boundary corresponds to a Transition, a Hard Cut, or a Sudden Jump.

\subsection{Automatic Video Transition Synthesis}

\begin{figure}[t]
    \centering
    \vspace{-0.3cm}
    \includegraphics[width=0.98\textwidth]{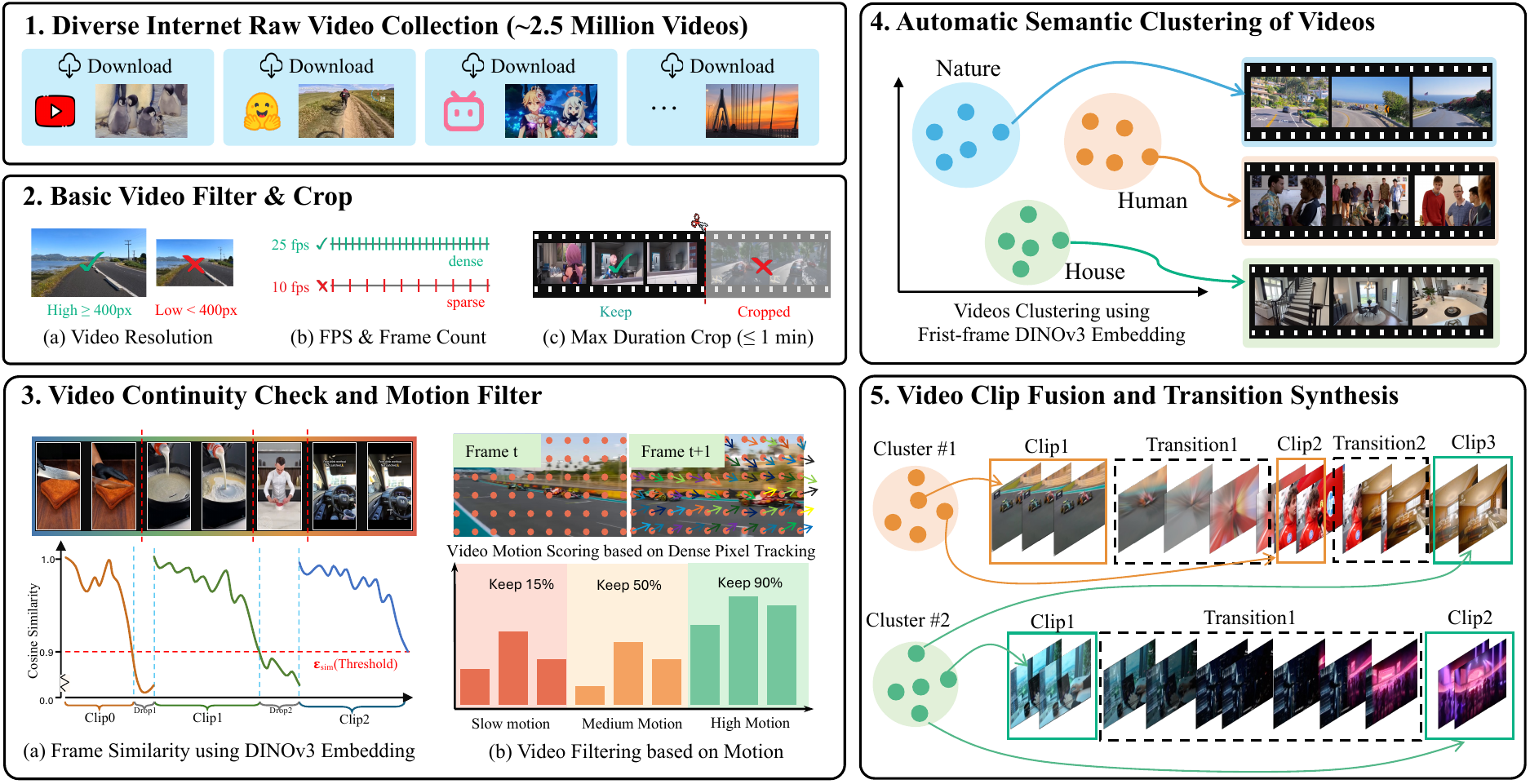}
    \caption{
    \textbf{Large-scale transition source video curation.}
    (1) We collect $\sim$2.5M raw videos from diverse Internet sources.
    (2) Videos are filtered based on resolution, frame rate, and duration constraints.
    (3) Temporal continuity and motion strength are verified using frame-level semantic similarity and dense motion tracking.
    (4) Remaining videos are automatically clustered using the SSL data curation method~\cite{vo2024automatic} to group semantically similar videos.
    (5) Finally, video clips in the same and different clusters are fused to synthesize large-scale shot boundary detection training datasets.
    }
    
    \label{fig:dataset_curation}
\end{figure}

To synthetically construct shot transitions, a clean video source pool is crucial. Our curation pipeline is shown in Fig.~\ref{fig:dataset_curation}. 
We first apply basic parameter filtering on duration, resolution, frames-per-second (fps), and aspect ratio to all collected in-the-wild video sources and crop them into segments with a maximum duration of 1 minute. 
To conservatively extract fluent video segments that contain no abrupt content change, inspired by motion evaluation of VBench~\cite{huang2024vbench}, we encode frames sampled at a constant interval into DINO~\cite{simeoni2025dinov3} embeddings, and compute the cosine similarity between consecutive embeddings. 
If the cosine similarity is higher than the threshold $\varepsilon_{sim}$, this indicates that the two frames are semantically consistent and do not exhibit an abrupt hard cut or transitions changing. 
We continue this process until the similarity below $\varepsilon_{sim}$, at which point the clip is terminated. And then we refresh the cache and start finding the next video clip.
Empirically, we observe that this approach is effective at identifying fading dark frames and dissolve-induced transparent frames, cropping out early when transitions start happening. 
Although this approach cannot guarantee that every extracted clip is completely clean, in subsequent synthetic transition generation, training with a large proportion of correctly labeled data can effectively mitigate the influence of noisy labels.

In this paper, identifying the \textit{Sudden Jump} is a critical task. 
Sudden Jump typically arise during video editing where a short period of segments is manually cropped, resulting in abrupt discontinuities, where the video cannot be regarded as a fluent shot.
Thus, we believe that sudden jump is aligned with the purpose of the shot boundary detection task and should be solved in this domain.
To incorporate into our shot boundary detection framework, we explicitly estimate the motion strength during the data curation stage. 
This enables us to select clips with medium-level motion intensity (neither too fast nor too slow) as suitable sources for constructing sudden jump samples.
We estimate motion strength using the CoTracker3~\cite{karaev2025cotracker3} model, which provides dense tracking points with configurable grid density. 
By measuring the displacement magnitude of tracked points across frames and averaging over all frames, we obtain an overall motion strength score for each video. 
Further, another functionality of motion strength information is that it can be helpful to increase video clip pool complexity.
This is because we observe that a large portion of raw clips exhibit small motion magnitudes; therefore, we filter out these slow-motion cases with weak dynamic patterns. 

Once each video clip is properly curated, we want to group \textbf{similar but not identical} videos into the same cluster, such that the sources before and after a synthetic transition can be sampled with high similarity to better simulate real-world video. 
We adopt the DINO representation and perform Self-Supervised Learning-based (SSL) clustering following the methodology of~\cite{vo2024automatic}. 
For each video clip, we extract the DINO embedding of its first frame, directly reusing the embeddings computed during the earlier curation stage. 
And then, we apply a semantic deduplication~\cite{abbas2023semdedup} paradigm to filter instances whose cosine similarity is less than the threshold $\varepsilon_{dup}$ in each cluster. This avoids near-duplicate videos when we are using large-scale random videos from the internet. 
Finally, we apply hierarchical K-means~\cite{vo2024automatic} clustering to group semantically similar embeddings. 
Each cluster represents a collection of videos sharing similar semantic content, like indoor scenes, vehicles, housing, mountains, etc. The SSL method does not directly tell us what the content is, but it can ensure that the contents are perceptually similar and strongly related. 

\paragraph{Synthetic Transition Composition.}
After the curation stage, we obtain a large collection of clean video clips. We then randomly choose videos from this pool and stitch clips together with diverse synthetic transitions (as shown in Fig.~\ref{fig:transition_types} and Fig.~\ref{fig:category}). 
In the synthesis, most video clips are sampled from the same SSL-grouped cluster pools as previous videos, while we also allow cross-cluster selection to reflect the unpredictability of real-world video editing (see Fig.~\ref{fig:dataset_curation} (5)). 
For sudden jump cases, we restrict the video source selection to middle motion strength. 
Excessive camera or object motion often induces large structural changes, making it difficult to distinguish from hard cuts, whereas some small motion cases (e.g., static talk-show scenarios) yield changes that are barely perceptible even to humans.
The synthesis settings are detailed in the supplementary.



\subsection{Shot Query-based Dense Video Transformer}

\begin{figure}[t]
    \centering
    \vspace{-0.3cm}
    \includegraphics[width=1.0\textwidth]{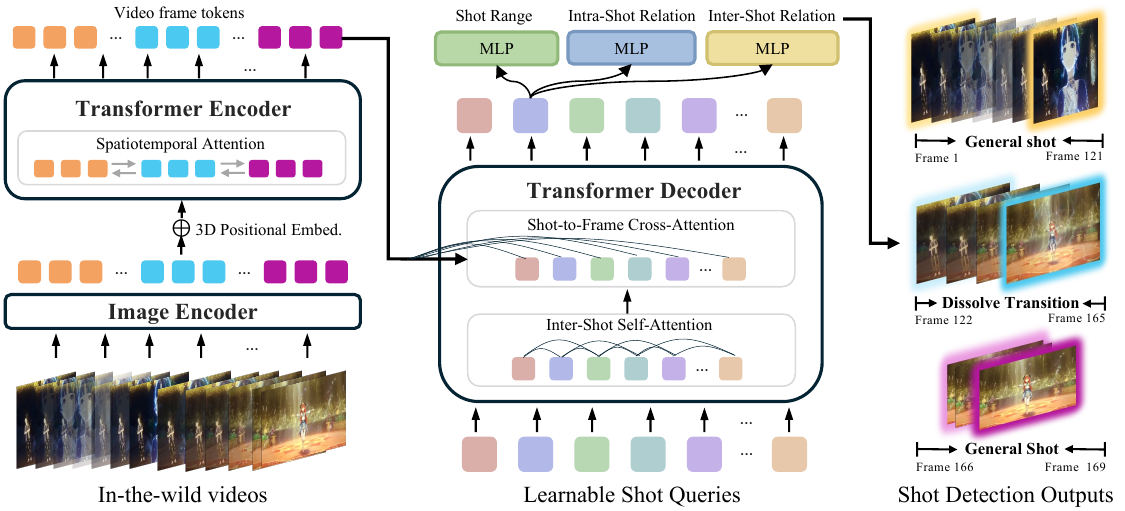}
    \caption{
        \textbf{Shot Query-based Dense Video Transformer.}
        Frame tokens from input videos are encoded using a spatiotemporal Transformer encoder with 3D positional embedding.
        Learnable shot queries in the decoder interact with frame features through cross-attention to predict shot range, intra-shot relation, and inter-shot relation.
    }
    \label{fig:transformer_architecture}
\end{figure}

As shown in Fig.~\ref{fig:transformer_architecture}, we propose a Shot Query-based end-to-end video Transformer model, which is composed of the image encoder, Transformer encoder, and Transformer decoder. 
We start from the image-based DETR~\cite{carion2020end} object detection Transformer model and introduce critical modifications for our task. 
The input consists of video frames of length $F$, height $H$, and width $W$, forming a tensor in $\mathbb{R}^{F \times H \times W \times C}$. 
The video is first encoded by ResNet~\cite{he2016deep} as an image encoder in a frame-by-frame manner. 

The encoded per-frame features are fed into the Transformer encoder. 
We flatten the spatial and temporal dimensions into a single dimension, resulting in a feature map of size $\mathbb{R}^ {d \times (F \cdot H \cdot W)}$, where $d$ is the hidden state dimension. 
Each Transformer encoder layer is composed of multi-head self-attention.
Since our input shifts from images to videos, the tokens in the Transformer remain permutation-invariant by design, we need to extend the 2D position embedding to 3D position embedding, introducing additional positional information along the temporal dimension. 
Specifically, following the approach of VisTR~\cite{wang2021end}, we generalize the cumulative spatial coordinates (x, y) to 3D (t, x, y) and apply sinusoidal embeddings along the temporal and spatial axes, enabling the Transformer to model joint spatiotemporal relationships in video inputs. The 3D position embedding will also be flattened to $\mathbb{R}^ {d \times (F \cdot H \cdot W)}$ and then added with the flattened video tokens before entering the Transformer encoder.

The input to the Transformer decoder is a fixed-length set of trainable embeddings, referred to as \emph{shot queries}. 
At a high level, each shot query serves as a shot prediction slot, aggregating shot-specific evidence from the video sources into a compact hidden state for decoding.
The entire Transformer decoder consists of multiple decoder layers. 
In each layer, the shot queries first undergo self-attention, followed by cross-attention with the tokens $\mathbb{R}^ {d \times (F \cdot H \cdot W)}$ produced by the Transformer encoder.
The number of input shot queries is fixed. 
At the output stage, shot queries will predict a dedicated termination token to explicitly indicate the end of shot prediction when it reaches the last shot.
All queries after the termination token are discarded, and only the preceding ones are considered valid predictions.

Each shot query on the output of our Transformer decoder is passed through three heads: a range head, an intra-relation head, and an inter-relation head. 
Directly adopting DETR-style~\cite{lei2021detecting} formulation by replacing bounding box prediction with an $L_1$ + 1D GIoU regression loss results in a suboptimal learning objective for temporal range prediction (check Sec.~\ref{sec:ablation}).
Regression over normalized continuous coordinates is inherently ill-suited for accurate frame-level boundary localization across long sequences, where even a one-frame deviation at a hard-cut transition constitutes a significant error for the SBD task.
To address this limitation, we reformulate range prediction as a discrete classification problem over frame indices, which provides improved localization precision and more stable optimization.
Specifically, the range head predicts the index of the last frame of each shot $p^{\text{end}}$, formulated as a classification problem where the number of classes equals the total number of frames.
As shots in SBD are consecutive and non-overlapping, the start of each shot is implicitly defined by the end of the previous one, with the first shot starting at frame 0.
This classification formulation for the range prediction does not require post-processing of heuristic thresholding like prior SBD methods~\cite{soucek2024transnet, zhu2023autoshot}.
Consequently, Hungarian matching~\cite{kuhn1955hungarian} is no longer required.
We retain auxiliary supervision at intermediate decoder layers to facilitate stable optimization.


\subsection{Evaluation Benchmark}

\begin{figure}[t]
    \centering
    \includegraphics[width=1.0\textwidth]{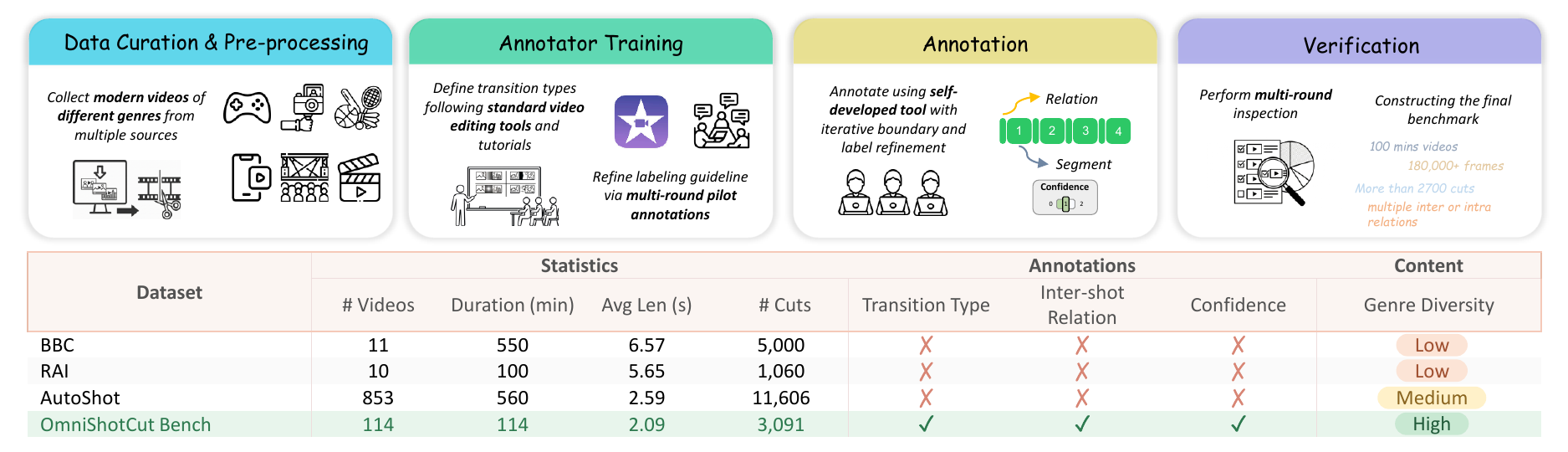}
    \caption{
    OmniShotCutBench Pipeline and Data Statistics.
    }
    \label{fig:benchmark_curation}
\end{figure}

We introduce OmniShotCutBench, a modern shot boundary detection benchmark designed to comprehensively evaluate models’ performance on versatile transitions from modern internet video sources. 
Each range label is paired with a confidence score.
This is because we recognize that human perception is inherently insensitive to subtle transition variations, particularly in transparent dissolve and fading effects.
OmniShotCutBench is constructed entirely from real edited videos and is disjoint from all synthetic training sources. 
No synthetic transitions generated by our pipeline are included in the benchmark.
The construction pipeline and statistics comparison is shown in Fig.~\ref{fig:benchmark_curation}.

We collect diverse video sources with modern video editing techniques from the topics of vlog, anime, movie, concert, documentary, monitor recording, game, sports, etc. 
We randomly truncate the videos to one minute (or shorter) and standardize all videos to 480p resolution at 30 FPS to ensure consistent temporal precision, which is critical for accurate shot boundary localization.
In total, we curate 114 videos, which is 114 minutes of diverse, high-quality, and representative video sources.

To ensure high-quality annotations, following other datasets \& benchmark works~\cite{lin2025towards, liu2025shotbench}, we mimic their high-standard curation paradigm, as shown in Fig.~\ref{fig:benchmark_curation}.
All annotators first studied several professional video editing tutorials online and learning video editing applications like iMovie. 
Annotators were required to review these materials prior to labeling to establish a clear understanding of transition taxonomy and visual characteristics.
We then conducted multiple rounds of pilot annotations to align labeling criteria and ensure consistency across annotators. 
During the final annotation phase, ambiguous or contentious cases were systematically documented and resolved to maintain annotation quality and consistency.
Analysis and visualization of our benchmark are provided in the supplementary material.
We will open-source this benchmark for future research.

%% file: sec/4_experiment.tex
\section{Experiment}

\subsection{Implementation Details}

We train our model with 8 Nvidia A100 GPUs for 50 epochs. 
Since training resolution is only 128x96, we choose the smallest pretrained ResNet18~\cite{he2016deep} as the image encoder to maintain more spatial tokens after the encoder.
We set 3 Transformer encoders and 6 Transformer decoders for our model, which is 34.49M parameters, out of which 11.17M is the ResNet backbone.
We set 24 fixed learnable shot query tokens as the input to the Transformer decoder, corresponding to the maximum number of shots observed in a 100-frame training clip.
The learning rate for the ResNet backbone is set to $1e-5$, and the Transformer encoder and decoder learning rate is $1e-4$.
Total batch size across all GPUs is 64.
We randomly crop 100 frames of the full video source for the training.
In the training, we do several online augmentations. 
This includes horizontal and vertical flip, color jittering, blurring, Gaussian and Poisson noises, and compression artifacts~\cite{wang2024vcisr}.
More implementation details are provided in the supplementary.

\subsection{Experiment Results}

Though our model jointly outputs the relational label, our capability is mainly inherited from shot boundary detection.
To evaluate traditional shot boundary detection, we choose to compare with mainstream baselines in the literature, which include non-learning-based method PySceneDetect~\cite{pyscenedetect2025}, and previous state-of-the-art learning-based methods by 3D CNNs, TransNet V2~\cite{soucek2024transnet}, and AutoShot~\cite{zhu2023autoshot}.

The evaluation is done on OmniShotCutBench and legacy benchmark BBC Planet Earth Documentation~\cite{baraldi2015deep}.
Our evaluation considers traditional shot range precision, recall, and F1 metrics, which are based on the ShotBench in Cosmos~\cite{agarwal2025cosmos}. We use their default tolerance of 2 frames.
Further, we specialize in the analysis of the transition of the IoU and sudden jump accuracy.
For transition IoU, we select the GT shots that are labeled with a transition label, and find the closest prediction results to calculate the IoU, applying our human label confidence to dynamically adjust the tolerance range.
For sudden jump accuracy, we identify all ground-truth inter-relation labels corresponding to sudden jumps and measure the proportion of correctly predicted shot cuts at the same frame index. A zero tolerance is applied, as the transition is expected to occur instantaneously.
We further evaluate intra- and inter-relation classification accuracy, which is the number of correct classifications divided by the total number of shots. For each ground-truth shot, the predicted shot with the highest IoU is selected and used for classification comparison.
All metrics count across all videos in the benchmark, instead of the average per video. This is because the number of cuts is unbalanced in different videos.







\begin{table}[t]
\caption{
    Quantitative comparison with existing shot boundary detection methods.
    We report results on the OmniShotCut Bench and the legacy BBC benchmark~\cite{baraldi2015deep}.
    For our benchmark, we additionally evaluate transition localization, sudden jump detection, and relation classification.
    All metrics are higher the better. The best is \textbf{highlighted}.
}
\centering
\small

\setlength{\tabcolsep}{3.5pt}
\renewcommand{\arraystretch}{1.02}

\begin{tabular}{lccccccc !{\vrule width 1.1pt} ccc}
\toprule
& \multicolumn{7}{c}{OmniShotCut Bench}
& \multicolumn{3}{c}{BBC Bench} \\
\cmidrule(lr){2-8}
\cmidrule(lr){9-11}

Method
& \makecell{Trans.\\IoU}
& \makecell{Sudden\\Jump\\Acc.}
& \multicolumn{3}{c}{Shot Range}
& \multicolumn{2}{c}{Relation Acc.}
& \multicolumn{3}{c}{Shot Range} \\[-0.55em]

\cmidrule(lr){4-6}
\cmidrule(lr){7-8}
\cmidrule(lr){9-11}

&
&
& Precision & Recall & F1
& Intra & Inter
& Precision & Recall & F1 \\[-0.15em]

\midrule

PySceneDetect
& 0.183 & 0.417 & 0.833 & 0.689 & 0.755 & - & -
& 0.893 & 0.884 & 0.889 \\

TransNet V2
& 0.193 & 0.262 & \textbf{0.913} & 0.735 & 0.814 & - & -
& 0.983 & 0.951 & 0.967 \\

AutoShot
& 0.253 & 0.455 & 0.849 & 0.783 & 0.815 & - & -
& \textbf{0.984} & 0.922 & 0.952 \\

\textbf{Ours}
& \textbf{0.644} & \textbf{0.759} & 0.904 & \textbf{0.859} & \textbf{0.881}
& \textbf{0.962} & \textbf{0.827}
& 0.967 & \textbf{0.974} & \textbf{0.971} \\

\bottomrule
\end{tabular}
\vspace{-0.3cm}
\label{tab:main_result}
\end{table}

Tab.~\ref{tab:main_result} reports the quantitative results. 
Traditional shot boundary detection methods such as PySceneDetect~\cite{pyscenedetect2025}, TransNetV2~\cite{soucek2024transnet}, and AutoShot~\cite{zhu2023autoshot} achieve reasonable performance on overall range-based metrics, with F1 scores between 0.75 and 0.82. 
However, they exhibit clear limitations in transition localization and sudden jump detection.
In particular, transition IoU remains low (0.18–0.25), indicating that predicted boundaries are often only roughly aligned with the true transition ranges. 
Sudden jump accuracy is also limited, suggesting difficulty in reliably detecting instantaneous discontinuities.
In contrast, our method significantly improves transition localization, achieving a transition IoU of 0.644, substantially outperforming all baselines. 
It also achieves the best range F1 score of 0.881. 
Moreover, our framework enables structured relation prediction, reaching 0.962 intra-shot accuracy and 0.827 inter-shot accuracy, which are not supported by prior methods.
In the legacy BBC benchmark~\cite{baraldi2015deep} comparison, we applied an overlap window of 10 between processing each clip and merging transition labels to match traditional hard-cut only output, and we achieved the best range recall and F1.
The visual result is available in the supplementary.

\subsection{Ablation Study}
\label{sec:ablation}

\begin{table*}[t]
\centering
\footnotesize
\setlength{\tabcolsep}{4pt}
\renewcommand{\arraystretch}{1.12}


\caption{
Ablation studies on key design choices. The best result within each ablation group is \textbf{highlighted} in bold.
}
\vspace{0.2em}
\label{tab:ablation_studies}

\begin{minipage}[t]{0.48\linewidth}
\centering
\textbf{(a) Loss Design}
\vspace{0.15em}

\begin{tabular*}{\linewidth}{@{\extracolsep{\fill}}lcccc}
\toprule
Loss & \makecell{Trans.\\IoU} & \makecell{Sudden\\ Jump\\Acc.} & \makecell{Range\\F1} & \makecell{Intra / Inter\\Acc.} \\
\midrule
Classification  
& 0.644 
& \textbf{0.759} 
& \textbf{0.881} 
& \textbf{0.962} / \textbf{0.827} \\
$L_1$ + 1D GIoU 
& \textbf{0.718} 
& 0.717 
& 0.857 
& 0.956 / 0.809 \\
\bottomrule
\end{tabular*}
\end{minipage}
\hfill
\begin{minipage}[t]{0.48\linewidth}
\centering
\textbf{(b) SSL Data Clustering}
\vspace{0.15em}

\begin{tabular*}{\linewidth}{@{\extracolsep{\fill}}lcccc}
\toprule
Setting & \makecell{Trans.\\IoU} & \makecell{Sudden\\Jump\\Acc.} & \makecell{Range\\F1} & \makecell{Intra / Inter\\Acc.} \\
\midrule
With 
& \textbf{0.644} 
& \textbf{0.759} 
& \textbf{0.881} 
& \textbf{0.962} / \textbf{0.827} \\
Without 
& 0.551 
& 0.664 
& 0.861 
& 0.957 / 0.722 \\
\bottomrule
\end{tabular*}
\end{minipage}

\vspace{0.85em}

\begin{minipage}[t]{0.48\linewidth}
\centering
\textbf{(c) Short Dense Cuts Distribution}
\vspace{0.15em}

\begin{tabular*}{\linewidth}{@{\extracolsep{\fill}}lcccc}
\toprule
Setting & \makecell{Trans.\\IoU} & \makecell{Sudden Jump\\Acc.} & \makecell{Range\\F1} & \makecell{Intra / Inter\\Acc.} \\
\midrule
With 
& \textbf{0.644} 
& \textbf{0.759} 
& \textbf{0.881} 
& \textbf{0.962} / \textbf{0.827} \\
Without 
& 0.610 
& 0.509 
& 0.831 
& 0.955 / 0.790 \\
\bottomrule
\end{tabular*}
\end{minipage}
\hfill
\begin{minipage}[t]{0.48\linewidth}
\centering
\textbf{(d) Encoder Quantity}
\vspace{0.15em}

\begin{tabular*}{\linewidth}{@{\extracolsep{\fill}}lcccc}
\toprule
Quantity & \makecell{Trans.\\IoU} & \makecell{Sudden Jump\\Acc.} & \makecell{Range\\F1} & \makecell{Intra / Inter\\Acc.} \\
\midrule
3 enc. 
& \textbf{0.644} 
& 0.759 
& \textbf{0.881} 
& \textbf{0.962} / \textbf{0.827} \\
6 enc. 
& 0.598 
& \textbf{0.774} 
& 0.879 
& 0.955 / 0.815 \\
\bottomrule
\end{tabular*}
\end{minipage}

\vspace{0.85em}

\begin{minipage}[t]{0.48\linewidth}
\centering
\textbf{(e) Training Resolution}
\vspace{0.15em}

\begin{tabular*}{\linewidth}{@{\extracolsep{\fill}}lcccc}
\toprule
Resolution & \makecell{Trans.\\IoU} & \makecell{Sudden Jump\\Acc.} & \makecell{Range\\F1} & \makecell{Intra / Inter\\Acc.} \\
\midrule
64$\times$96 
& 0.582 
& 0.699 
& 0.877 
& 0.959 / 0.818 \\
96$\times$128 
& \textbf{0.644} 
& \textbf{0.759} 
& \textbf{0.881} 
& \textbf{0.962} / \textbf{0.827} \\
128$\times$160 
& 0.584 
& 0.723 
& 0.867 
& 0.954 / 0.794 \\
\bottomrule
\end{tabular*}
\end{minipage}
\hfill
\begin{minipage}[t]{0.48\linewidth}
\centering
\textbf{(f) Number of Shot Queries}
\vspace{0.15em}

\begin{tabular*}{\linewidth}{@{\extracolsep{\fill}}lcccc}
\toprule
Number & \makecell{Trans.\\IoU} & \makecell{Sudden Jump\\Acc.} & \makecell{Range\\F1} & \makecell{Intra / Inter\\Acc.} \\
\midrule
12
& 0.556
& 0.735
& 0.876
& 0.953 / \textbf{0.833} \\
24 
& \textbf{0.644}
& \textbf{0.759} 
& \textbf{0.881} 
& \textbf{0.962} / 0.827 \\
48 
& 0.590 
& 0.753
& 0.873 
& 0.959 / 0.811 \\
\bottomrule
\end{tabular*}
\end{minipage}
\label{tab:ablation}
\end{table*}

In this section, we conduct ablation studies to examine the impact of key components and hyperparameter influence on our performance.
Results are summarized in Tab.~\ref{tab:ablation}.

For the first study, we examine whether the DETR-style range regression objective, $L_1$ + 1D GIoU~\cite{carion2020end}, is preferable to our proposed formulation, which converts boundary estimation to discrete classification.
In our model, we directly predict discrete boundary labels like classification.
For the $L_1$ + 1D GIoU variant, the model outputs are passed through a sigmoid to map into $[0,1]$ range, and then scaled by the maximum prediction length (100 frames in our training). Default 2D GIoU is changed to a 1D version, based on~\cite{lei2021detecting}.
As shown in Tab.~\ref{tab:ablation} (a), $L_1$ + 1D GIoU can slightly improve transition IoU, but it degrades under stricter criteria, such as zero-tolerance sudden jump accuracy and range F1.
We attribute this drop to the inherent difficulty of regression losses in resolving the last 1-2 frames precisely,
which is crucial for abrupt hard-cut and sudden-jump boundaries.

In Tab.~\ref{tab:ablation} (b), we study the influence of sampling transition sources from the same SSL-curated clusters~\cite{simeoni2025dinov3, vo2024automatic} versus purely random selection.
In our base setting, we follow~\cite{vo2024automatic} and sample transition sources from the same cluster with $75\%$ probability.
Hereby, we construct a variant where all clips are selected at random.
\emph{Pure random sampling consistently degrades performance across all metrics.}
We hypothesize that semantically aligned clips produce more challenging synthesized transitions during training, requiring the model to leverage fine-grained temporal and structural cues to identify subtle content changes.
In contrast, randomly paired clips often exhibit large semantic discrepancies, making the discrimination task comparatively trivial.

Next, we aim to study whether modeling continuous dense hard cuts and the sudden jump during the data synthesis is beneficial. 
Under purely random synthesis, fewer than 0.005\% of generated samples contain more than five consecutive hard cuts, despite such patterns being common in real-world videos.
Therefore, in our base setting, we enforce continuous hard cuts in $25\%$ of the synthesized videos.
Removing this strategy leads to consistent performance degradation across all metrics, as shown in Tab.~\ref{tab:ablation} (c).
These results suggest that understanding real-world shot distributions and designing synthesis pipelines that better reflect such distributions are important for effective model training.


For the architectural ablations in Tab.~\ref{tab:ablation} (d)--(f), the base configuration provides the most balanced overall performance.
Increasing the number of encoder layers improves sudden jump accuracy, but leads to noticeable degradation in most other metrics. 
A training resolution of 96$\times$128 gives the best resolution trade-off.
For the number of shot queries, using 24 queries yields the strongest overall results, achieving the best sudden jump accuracy and range F1, whereas 12 and 48 queries only improve transition IoU and inter-label accuracy, respectively.


%% file: sec/5_conclusion.tex
\section{Conclusion}

In this paper, we present \textbf{OmniShotCut}, reformulating shot boundary detection with explicit \textit{intra}-shot and \textit{inter}-shot relations by a shot query-based Transformer framework. 
To overcome the limitations of manual annotation, we develop a fully synthetic data synthesis pipeline that automatically produces diverse transition effects with precise temporal supervision, and curate a modern complex shot boundary detection benchmark. 
Experiments demonstrate that our approach achieves superior performance across multiple benchmarks and evaluation metrics. 
Our results suggest that fully synthetic supervision provides a scalable and effective paradigm for next-generation shot boundary detection datasets. 
Moreover, our insights into intra- and inter-shot relations may further benefit downstream applications that require more accurate and explainable shot boundary detection.

\paragraph{Limitations and Future Work.}
Despite the strong performance, more sophisticated artistic and semantically dynamic transitions may require additional modeling beyond our current synthetic parameterization. 
In particular, capturing complex cinematic transition patterns could benefit from large-scale industry-level transition template collections, which are not publicly available. 
Exploring such resources remains an interesting direction for future work.

%% file: sec/6_supp.tex

\renewcommand{\thesection}{\Alph{section}}
\setcounter{section}{0}

\section*{Supplementary Material}

This supplementary material provides more implementation and technical details and additional qualitative visualization to complement the main manuscript. 
In Sec.~\ref{sec:transition_genre}, we present full transition genre types.
In Sec.~\ref{sec:transition_synthesis_detail}, we present more information about the transition synthesis parameter setting. 
In Sec.~\ref{sec:more_implementation_details}, we present more implementation details for the model training and setting. 
In Sec.~\ref{sec:benchmark_label}, we present the benchmark annotation GUI and details.
In Sec.~\ref{sec:visual_comparisons}, we present the visual comparisons between different models.

\section{Full Transition Genre Types}
\label{sec:transition_genre}

\begin{figure}[t]
    \centering
    \includegraphics[width=1.0\textwidth]{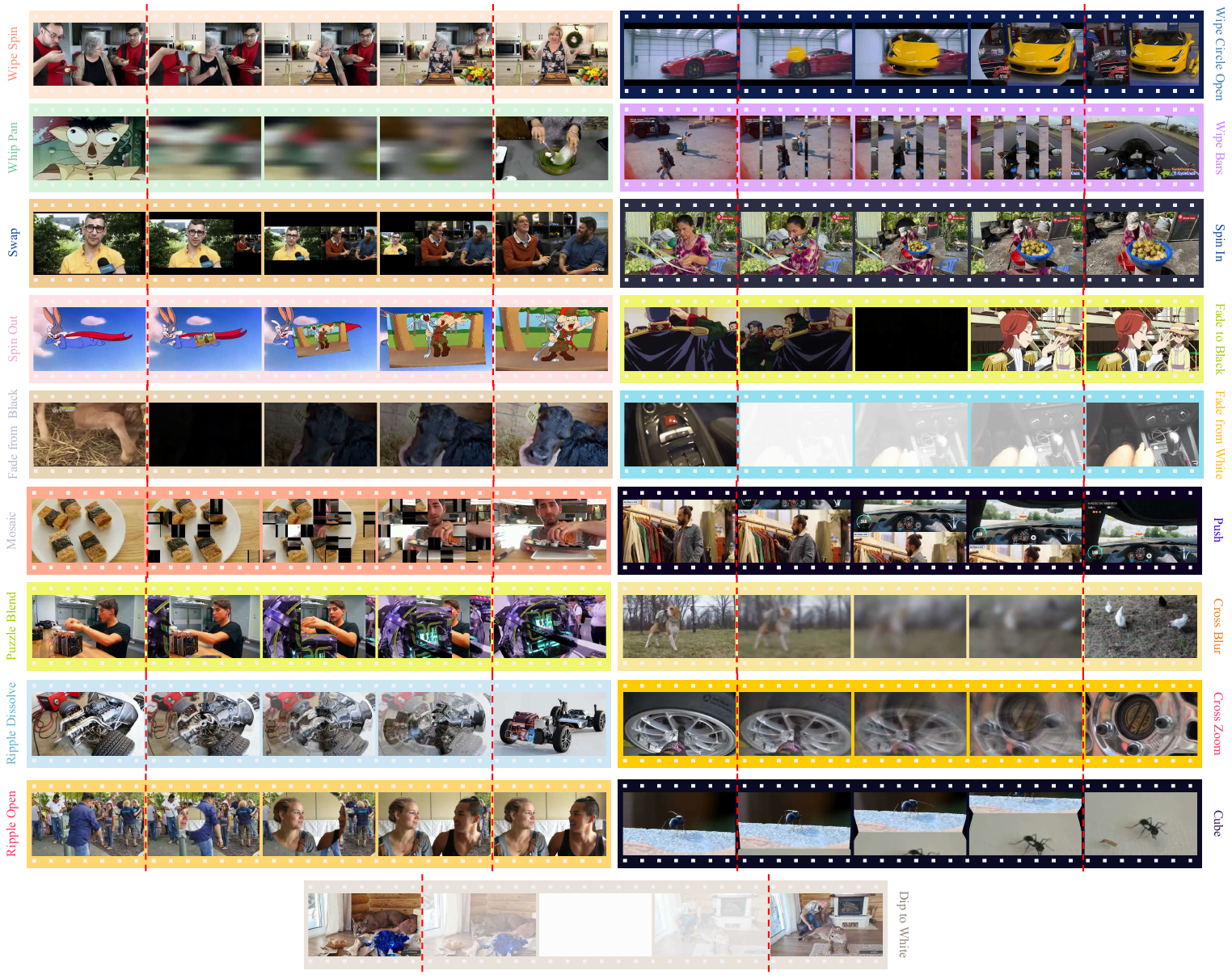}
    \vspace{-0.8cm}
    \caption{
        \textbf{Full Transition Types}.
        We complement the transition types listed in the main paper section.
    }
    \label{fig:extra_transition_types}
\end{figure}

In the main paper, we provide the main transition type visualization. However, this is still not all the types we consider. We consider numerous subtypes in transition and classify them into different categories based on the pattern. The visualization is shown in Fig.~\ref{fig:extra_transition_types}.

We consider a diverse taxonomy of editing transitions covering both common and fine-grained variants. 
Specifically, the transition set includes Dissolve transitions (Transparent Dissolve, Cross-Blur Dissolve, and Ripple Dissolve); 
Wipe transitions (Unidirectional Wipe, Diagonal Wipe, Circular Wipe, Bar Wipe, Ripple Wipe, Page-Curl Wipe, and Mosaic Wipe); 
Push transitions (Unidirectional Push and Puzzle Push); 
Slide transitions (Horizontal Slide, Whip-Pan Slide, and Cube Slide); 
Zoom transitions (Zoom In/Out, Spin In/Out, Cross Zoom, and Swap Zoom); 
Fade transitions (Fade to Black, Fade to White, Fade from Black, Fade from White, Dip to Black, and Dip to White); 
and Doorway transitions (Doorway Open).

\section{Transition Synthesis Details}
\label{sec:transition_synthesis_detail}

\begin{figure}[t]
    \centering
    \vspace{-0.2cm}
    \includegraphics[width=0.95\textwidth]{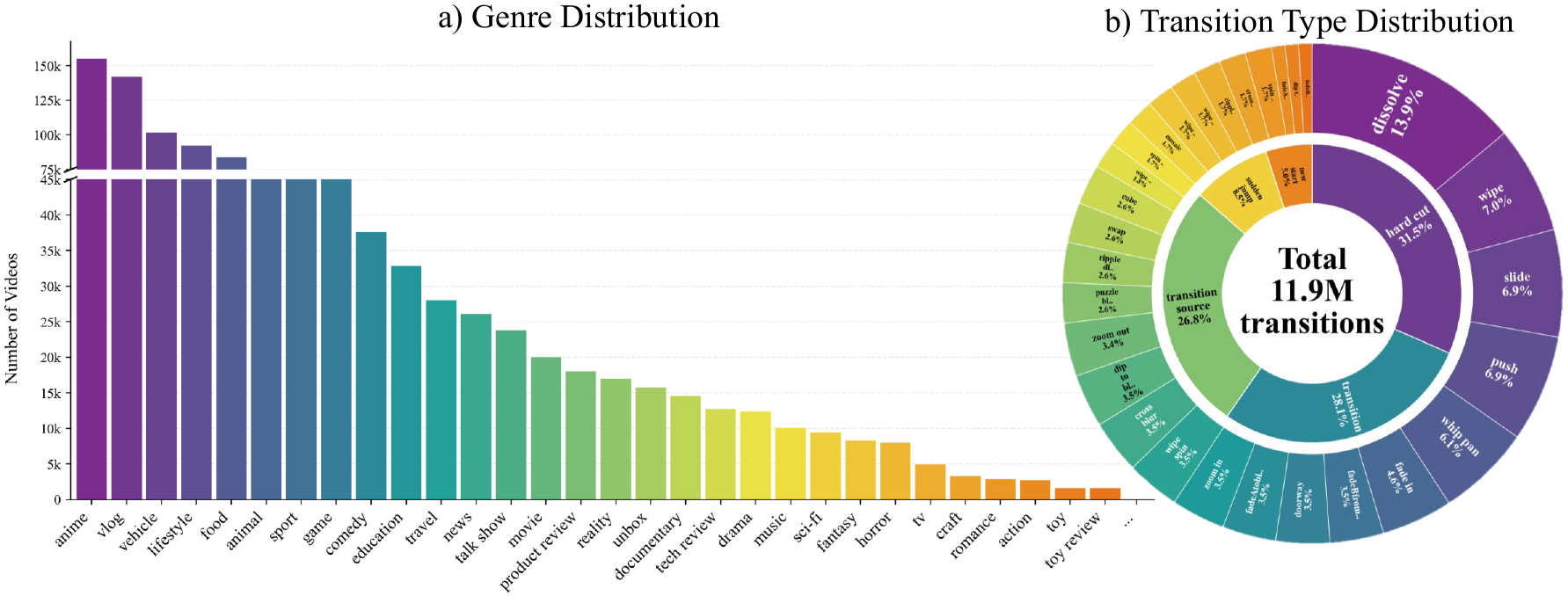}
    \vspace{-0.3cm}
    \caption{
        \textbf{Left:} Our curation pipeline scales to internet-scale, wide-domain video collection and yields $\sim$1.5M curated clip source for transition synthesis; genres are annotated by Qwen3~\cite{yang2025qwen3}.
        \textbf{Right:} Transition statistics of the synthetic corpus. The inner ring shows the inter-shot relation distribution, and the outer ring breaks down all of the main and sub-transition types that our pipeline can synthesize. In total, we synthesize 11.9M transitions for training.
    }
    \vspace{-0.2cm}
    \label{fig:category}
\end{figure}

In this section, we share our transition synthesis details, which are the key to our training dataset construction. A preview can be found in Fig.~\ref{fig:category}.

In the curation stage, our video sources are mostly coming from existing video datasets on Huggingface, which include datasets like OpenVid~\cite{nan2024openvid}, VidGen~\cite{tan2024vidgen}, Sakuga~\cite{pan2024sakuga}, GamePhysics~\cite{taesiri2022clip}, and several publicly accessible sources. 
we set continuous similarity threshold $\varepsilon_{sim}$ to 0.9 and deduplication threshold $\varepsilon_{dup}$ to 0.05. 
The motion tracking~\cite{karaev2025cotracker3} is sampled per 3 frames at 256x320 resolution.
The number of SSL clusters~\cite{vo2024automatic} is set to 27,000, where we use DINOv3~\cite{simeoni2025dinov3} ViT large variant.
We discard cluster sizes that are less than 5 videos to avoid reusing the same video source.

We construct our synthetic transition training data via a fully parameterized pipeline. 
The number of clips per video is sampled from a Poisson distribution with $\lambda=7.0$ and constrained to $[1,28]$.
Clip durations are sampled from Gaussian distributions of $\mathcal{N}(2.8,1.6^2)$ seconds. 
$75\%$ of clips are selected from the same DINOv3~\cite{simeoni2025dinov3} cluster to maintain semantic coherence.
For sudden-jump cases, we crop [24, 40] frames, and their valid source videos should be those with motion strength in the [25, 60] percentile range, sorted from slowest to fastest.
We further assign $25\%$ of the synthesis to be extremely short and dense composition, where we generate continuous 28 videos that are within the duration of $[0.15, 1.0]$ seconds for each clip. 
For the offline augmentation, we apply 5\% on adding subtitle text, and 7.5\% on the lighting variations.
In total, we create 300K synthetic videos used for training, where each of them contains at least 240 frames at 24 fps.
The quantity of synthesis videos can be infinite, but we set it to 300K videos as a reasonable range.


The distribution of each transition is different. We sample 35\% for the hard cut.
In the dissolve type transition, 9.4\% is distributed for the transparent dissolve, 2.4\% for the cross-blur dissolve, and 1.8\% for the ripple dissolve.
In the wipe type transition, 4.7\% is distributed evenly for the vanilla wipe in up, down, bottom, and right directions, 2.4\% for spin wipe, 2.4\% for circle open and close wipe, 1.2\% for the bar wipe, 1.2\% for the ripple wipe, and 1.2\% for the mosaic wipe.
In the push-type transition, 4.7\% is distributed evenly for the vanilla push in up, down, bottom, and right directions, and 1.8\% for the puzzle blending push.
In the slide-type transition, 4.7\% is distributed evenly for the vanilla slide, 4.1\% for the whip pan, and 1.8\% for the cube slide.
In the zoom-type transition, 2.4\% is distributed for the zoom in, 2.4\% for the zoom out, 2.4\% for the spin in and out, 1.2\% for the cross zoom, and 1.8\% for the swap.
In the fading-type transition, 2.9\% is distributed for fading the first source to black or white screen, 2.9\% for fading the second source from black or white screen, and 2.9\% for the dip fading effect.
In the doorway transition, 2.9\% is distributed.

For all transition types, we carefully control as many parameters as possible to ensure consistency and precise manipulation.
This yields a large set of explicit control philosophies spanning 
(i) discrete mode switches (e.g., transition direction, hard vs. soft edges, constant vs. linear smoothing), 
(ii) temporal controls (including the start time, duration, and the speed curve of the transition over time),
(iii) spatial controls (anchor locations and margins for added text, effect centers for zoom/ripple, grid resolution for mosaic, doorway seam orientation), 
and (iv) intensity controls (blurring range and curve shape, zoom magnitude and sampling density, lighting gains/gamma/contrast and color wash/spotlight strength, feather widths for soft boundaries). 
Additionally, we control content-level factors such as text selection and layout (wrapping, line count, spacing), while dropping near-duplicate frames on the edge of transition phases. 
Overall, our transition synthesis pipeline defines a reproducible distribution over diverse transitions with fine-grained, interpretable parameters that can be tuned.
To implement the transition, we apply LLM as an auxiliary tool, but humans do the final check to ensure that the transition is correct.

\section{More Implementation Details}
\label{sec:more_implementation_details}

For the model training, we optimize a weighted sum of three classification losses:
\begin{equation}
\mathcal{L}
=
\lambda_{\text{range}} \, \mathcal{L}_{\text{range}}
+
\lambda_{\text{intra}} \, \mathcal{L}_{\text{intra}}
+
\lambda_{\text{inter}} \, \mathcal{L}_{\text{inter}},
\end{equation}
where
{\setlength{\jot}{0pt}\abovedisplayskip=3pt\belowdisplayskip=3pt
\begin{align}
\mathcal{L}_{\text{range}}
&= \frac{1}{N} \sum_{i=1}^{N}
\mathrm{CE}\!\left(p^{\text{end}}_i,\; y^{\text{end}}_i\right), \\
\mathcal{L}_{\text{intra}}
&= \frac{1}{N} \sum_{i=1}^{N}
\mathrm{CE}\!\left(p^{\text{intra}}_i,\; y^{\text{intra}}_i\right), \\
\mathcal{L}_{\text{inter}}
&= \frac{1}{N} \sum_{i=1}^{N}
\mathrm{CE}\!\left(p^{\text{inter}}_i,\; y^{\text{inter}}_i\right).
\end{align}
}

In the experiment, $\lambda_{\text{range}}, \lambda_{\text{intra}}, \lambda_{\text{inter}}$ is set to 5, 1, 1.

\section{Benchmark Annotation Details}
\label{sec:benchmark_label}

\begin{figure}[t]
    \centering
    \includegraphics[width=1.0\textwidth]{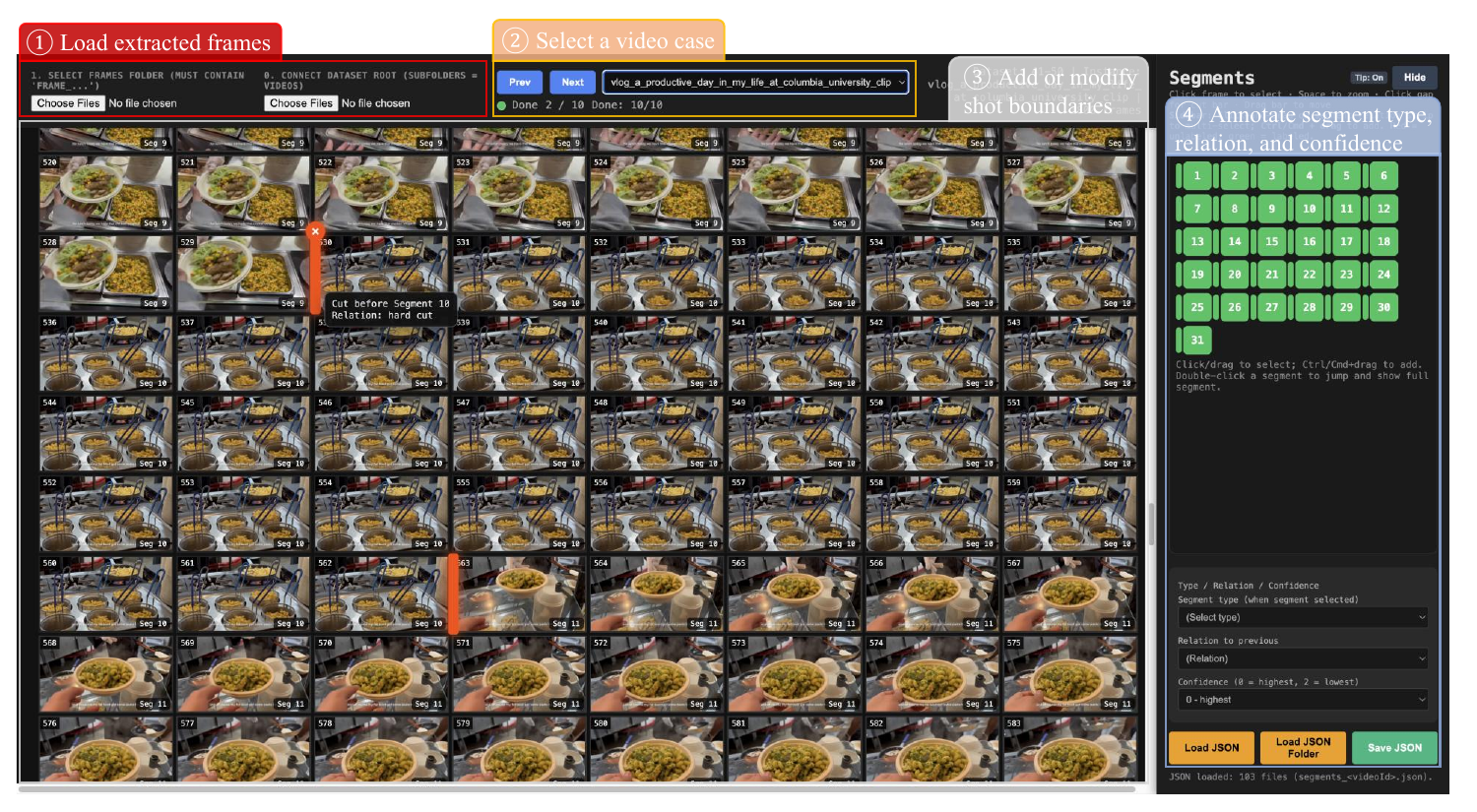}
    \caption{
    \textbf{Benchmark Annotation Tool}.
    Annotators first load extracted frames and select a video case. 
    Shot boundaries are created by clicking between frames along the timeline. 
    The right panel provides an overview of segments and enables labeling of type, relation, and confidence. Additional features, including multi-selection, auto-save, and frame-level inspection, facilitate efficient dataset construction.
    }
    \label{fig:label_tool}
\end{figure}

\begin{figure}[t]
    \centering
    \includegraphics[width=0.95\textwidth]{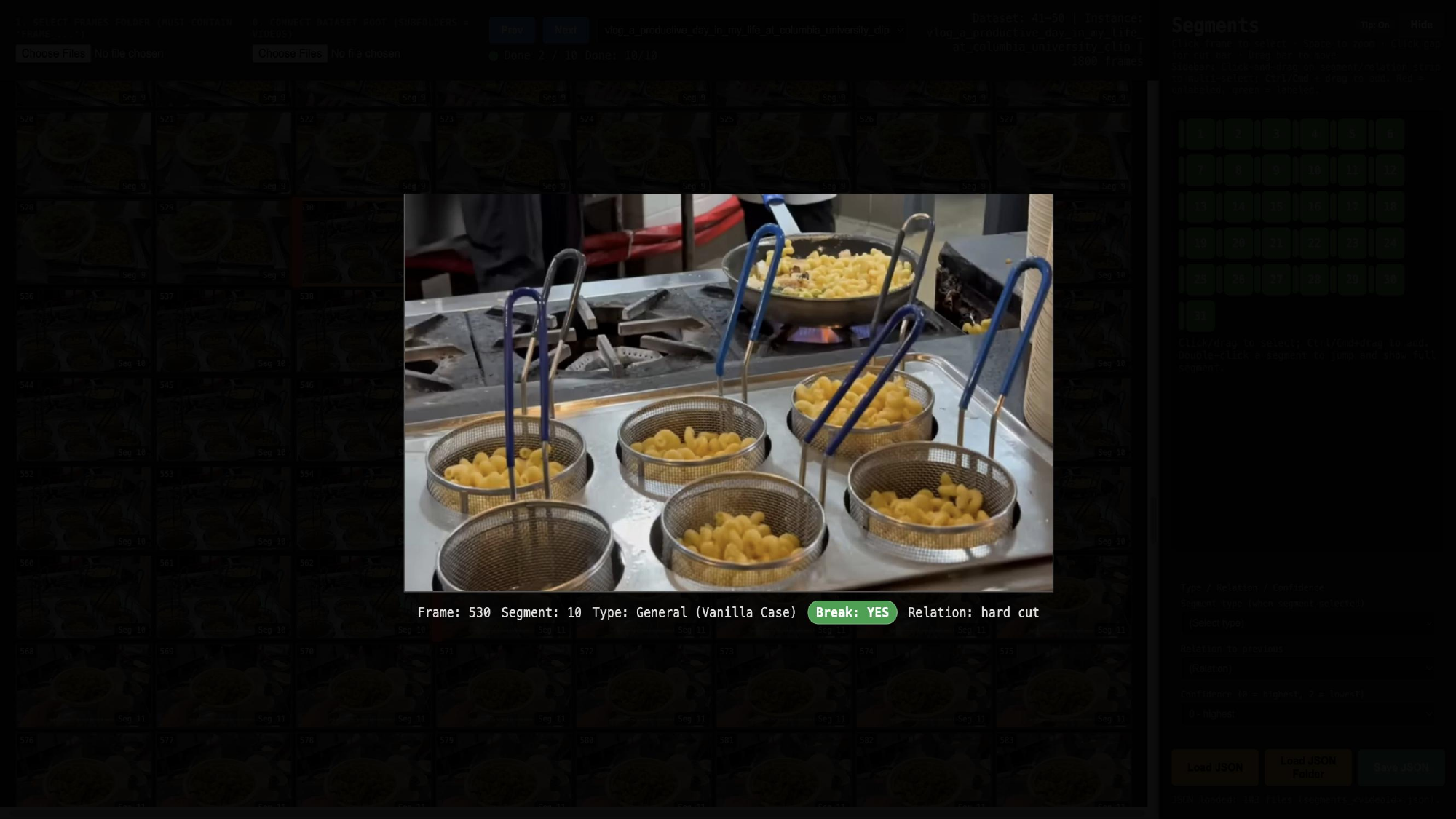}
    \caption{
    \textbf{Annotation Tool Open-Image Inspection Mode}. 
    The Inspection mode shows the high resolution and labeling details for the annotators to define subtle transition changes accurately. 
    It can be played as videos fluently by pressing the buttons to check gradual transitions frame-by-frame.
    }
    \label{fig:label_tool_zoom}
\end{figure}

As shown in Fig.~\ref{fig:label_tool}, we develop an annotation tool for our OmniShotCut bench.
This tool helps us swiftly locate the boundaries on the per-frame level and label dense transition cases for long video instances.
To facilitate efficient annotation, we implement several useful features. 
A floating window dynamically displays the current segment’s labels (type and confidence) or the relation label, allowing annotators to quickly verify existing annotations. 
In addition, the tool supports multi-selection, enabling annotators to label multiple segments or relations simultaneously with a single action. An auto-save mechanism is also integrated to automatically store labeling progress and prevent data loss.

\begin{figure}[t]
    \centering
    \includegraphics[width=1.0\textwidth]{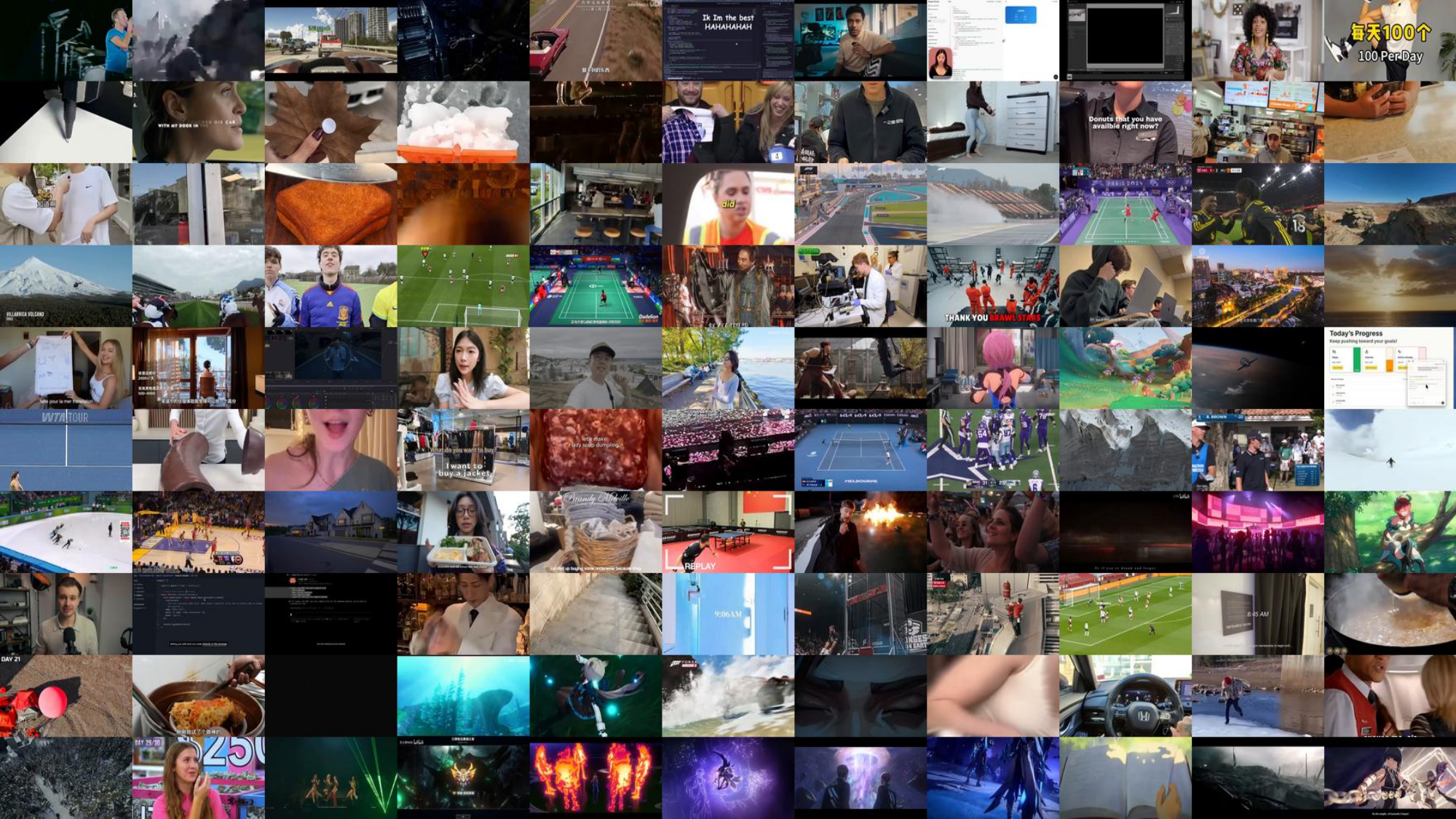}
    \caption{
        \textbf{OmniShotCut Bench Sample Images}.
        Our benchmark covers diverse topics spanning lifestyle, sports, entertainment, anime, game, unboxing, vlog, shorts, tutorials, urban scenes, screen-based media, and so on. 
    }
    \label{fig:all_benchmark}
\end{figure}

Certain transitions of interest in our benchmark, such as sudden jumps, dissolves, and fades, often involve subtle frame-level changes that require careful inspection. To address this, we introduce an open-image inspection mode (see Fig.~\ref{fig:label_tool_zoom}). Annotators can open a frame by double-clicking or pressing the space bar, which displays the frame along with its associated type, confidence, and relation annotations. 
Using the left and right arrow keys, annotators can navigate through frames sequentially, effectively previewing the sequence as a short video clip. These features together provide an efficient and user-friendly platform for constructing and verifying our benchmark dataset.
A preview of the videos in the benchmark is shown in Fig.~\ref{fig:all_benchmark}.

\section{Visual Comparisons}
\label{sec:visual_comparisons}

\begin{figure}[t]
    \centering
    \includegraphics[width=1.0\textwidth]{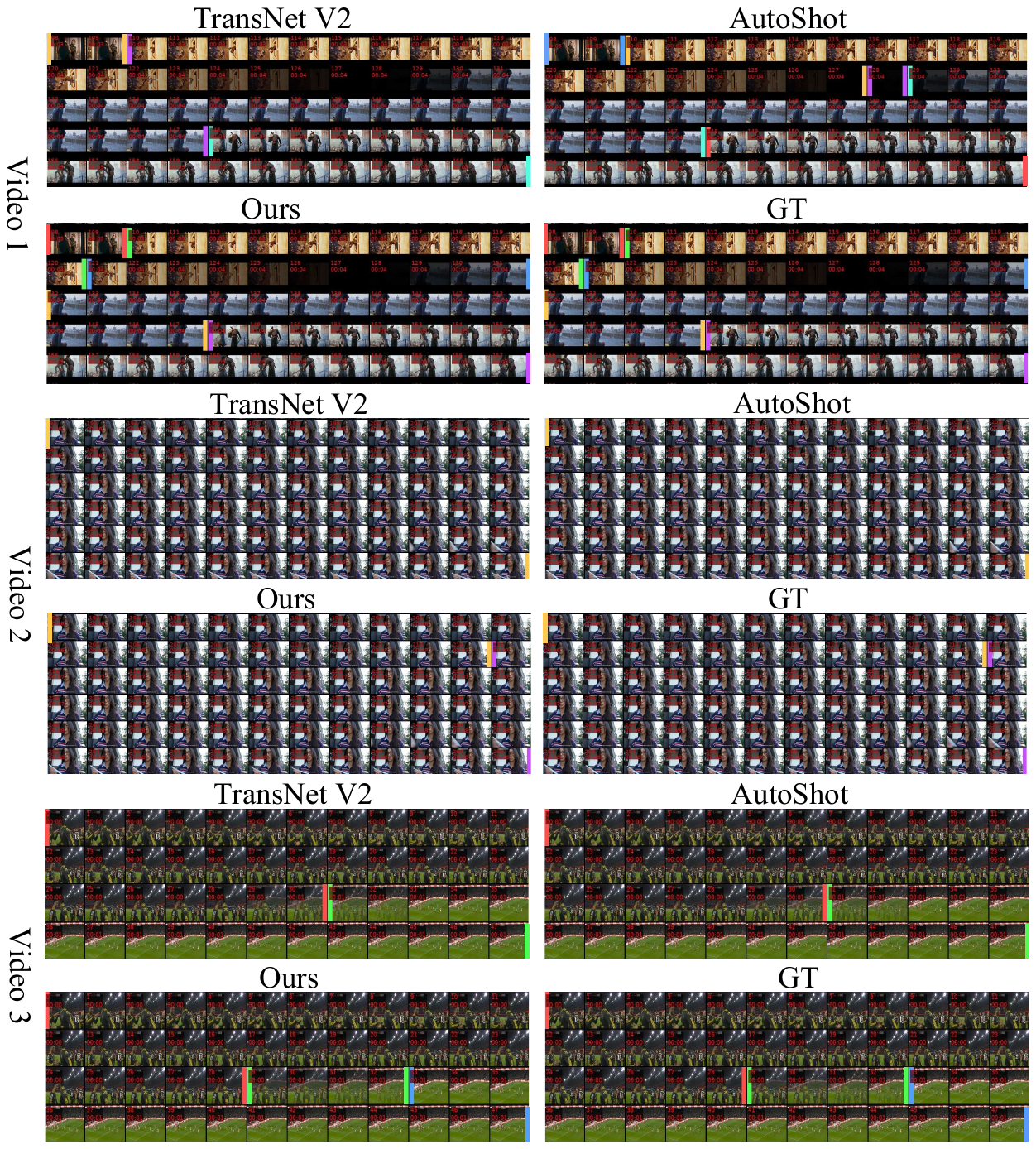}
    \caption{
        \textbf{Shot Boundary Detection Qualitative Comparisons}.
        We compare TransNet V2~\cite{soucek2024transnet}, AutoShot~\cite{zhu2023autoshot}, and ours on Fading (video 1), Sudden Jump (video 2), and Dissolve (video 3). 
        Each vertical bar with the same color denotes the start and end of a clip cut by the model.
        \textbf{Zoom in for the best view.}
    }
    \label{fig:qualitative_results}
\end{figure}

The visual comparison result is shown in Fig.~\ref{fig:qualitative_results}.
As we can see, our model succeeded in the fading and dissolve transition as well as sudden jump detection. 
Baseline models like TransNet V2~\cite{soucek2024transnet} and AutoShot~\cite{zhu2023autoshot} failed, where they usually choose the start frame in the middle of the dissolve or fading effects. 
This confusing first frame is not friendly for the downstream applications like video generation, where they demand a clear first frame source for the image-to-video generation.
Our result aligns with the ground-truth labels.
Further, these baselines cannot realize the sudden jump, which lacks sensitivity to the subtle changes.

